\documentclass[conference]{IEEEtran}
\IEEEoverridecommandlockouts
\usepackage{float}
\usepackage{amsmath}
\usepackage {diagbox}
\usepackage{url}
\usepackage{mathtools}
\usepackage{comment}
\usepackage{cite}
\usepackage{amsmath,amssymb,amsfonts}
\usepackage{algorithmic}
\usepackage{graphicx}
\usepackage{textcomp}
\usepackage{xcolor}
\usepackage{color}

\usepackage[whole]{bxcjkjatype} 
\usepackage[unicode]{hyperref}

\def\BibTeX{{\rm B\kern-.05em{\sc i\kern-.025em b}\kern-.08em
    T\kern-.1667em\lower.7ex\hbox{E}\kern-.125emX}}
\begin{document}

\title{ClaimBrush: A Novel Framework for Automated Patent Claim Refinement Based on Large Language Models}

\makeatletter
\newcommand{\linebreakand}{%
  \end{@IEEEauthorhalign}
  \hfill\mbox{}\par
  \mbox{}\hfill\begin{@IEEEauthorhalign}
}
\makeatother

\author{
\IEEEauthorblockN{Seiya Kawano$^{1,2}$, Hirofumi Nonaka$^{3}$, Koichiro Yoshino$^{4,1,2}$}
\IEEEauthorblockA{
$^{1}$Guardian Robot Project, RIKEN, Kyoto, Japan
$^{2}$Nara Institute of Science and Technology
Nara, Japan\\
$^{3}$Aichi Institute of Technology, Aichi, Japan
$^{4}$Tokyo Institute of Technology, Tokyo, Japan
}
\IEEEauthorblockA{
{seiya.kawano@riken.jp},{koichiro@c.titech.ac.jp},{hnonaka@aitech.ac.jp}
}

%\name{Kawano$^{1,2}$, Hirofumi Nonaka$^{3}$, Koichiro Yoshino$^{4,1,2}$}
%\address{$^{1}$Guardian Robot Project, RIKEN, $^{2}$Nara Institute of Science and Technology\\$^{3}$Aichi Institute of Technology, $^{4}$Tokyo Institute of Technology\\{seiya.kawano@riken.jp},{koichiro@c.titech.ac.jp},{hnonaka@aitech.ac.jp}}

%\IEEEauthorblockN{1\textsuperscript{st} Kiyotada Mori}
%\IEEEauthorblockA{Nara Institute of Science and Technology\\
%Nara, Japan\\
%mori.kiyotada.mh5@naist.ac.jp}
%\and
%\IEEEauthorblockN{2\textsuperscript{nd} Seiya Kawano}
%\IEEEauthorblockA{Guardian Robot Project, RIKEN\\
%Kyoto, Japan\\
%Nara Institute of Science and Technology\\
%Nara, Japan\\
%seiya.kawano@riken.jp}
%\and
%\IEEEauthorblockN{3\textsuperscript{rd} Chaoran Liu}
%\IEEEauthorblockA{Guardian Robot Project, RIKEN\\
%Kyoto, Japan\\
%chaoran.liu@riken.jp}
%\linebreakand
%\IEEEauthorblockN{4\textsuperscript{th} Carlos-Toshinori Ishi}
%\IEEEauthorblockA{Guardian Robot Project, RIKEN\\
%Kyoto, Japan\\
%carlos.ishi@riken.jp}
%\and
%\IEEEauthorblockN{5\textsuperscript{th} Koichiro Yoshino}
%\IEEEauthorblockA{Tokyo Institute of Technology\\
%Tokyo, Japan\\
%Guardian Robot Project, RIKEN\\
%Kyoto, Japan\\
%Nara Institute of Science and Technology\\
%Nara, Japan\\
%koichiro.yoshino@riken.jp}
}

\maketitle

\begin{abstract}
Automatic refinement of patent claims in patent applications is crucial from the perspective of intellectual property strategy. In this paper, we propose ``ClaimBrush,'' a novel framework for automated patent claim refinement that includes a dataset and a rewriting model. We constructed a dataset for training and evaluating patent claim rewriting models by collecting a large number of actual patent claim rewriting cases from the patent examination process. Using the constructed dataset, we built an automatic patent claim rewriting model by fine-tuning a large language model. Furthermore, we enhanced the performance of the automatic patent claim rewriting model by applying preference optimization based on a prediction model of patent examiners' Office Actions. The experimental results showed that our proposed rewriting model outperformed heuristic baselines and zero-shot learning in state-of-the-art large language models. Moreover, preference optimization based on patent examiners' preferences boosted the performance of patent claim refinement.
\end{abstract}

%%{keywords}

\section{Introduction}

Patent claims are part of a patent document that defines the technical scope protected by the patent, and patent claims consist of one or more claims~\cite{shinmori2003patent,marco2019patent}. In patent applications, it is important to effectively refine the content of patent claims in order for the invention to receive a patent grant or to obtain stronger rights compared to existing patents~\cite{faber1990landis,reilly2018amending}. With the increasing reliance on data-driven approaches in various industries, automating the refinement process has become a critical step in improving the efficiency of patent applications. However, refining patent claims is often a time-consuming and labor-intensive task, so effective tools to support the patent claim refinement are in demand~\cite{shinmori2008multi,tonguz2021automating}. 

% wipo 引用

While research has been conducted on basic patent information processing tasks such as patent retrieval and patent mining~\cite{lupu2017patent,lupu2017current,fujii2007overview}, as well as on supporting the understanding of patent claims~\cite{shinmori2003patent}, more work is needed on frameworks that suggest patent claim rewriting plans from various perspectives to provide more effective patent revision support. To realize such rewriting, a dataset containing rewriting examples from the viewpoints of what kinds of rewrites are effective in obtaining patent approval or strengthening the scope of rights is essential. This is because the state-of-the-art in rewriting research is based on statistical models such as language models, and data is indispensable for training and fine-tuning such models.

In this paper, we propose ``ClaimBrush,'' a novel framework that includes a rewriting model and a dataset for training and evaluation, with the goal of automatically generating rewrites of patent claims. We focus on the process before and after approval in patent examination and construct the dataset from the perspective of {\it{rewriting patent claims into patent claims that are more likely to be approved}}. Specifically, we focus on the examination process in patent applications and regard the pairs of claims described in the publication before examination and the patent publication tied to a particular patent application as examples of {\it{desirable}} rewrites of patent claims. These rewriting examples reflect the results of all amendments (both voluntary amendments and amendments in response to Office Actions\footnote{{\footnotesize An Office Action is a document sent by a patent examiner to an applicant, which presents the examiner's objections, rejections, and/or requirements concerning the patent application}}) made by the applicant during the patent examination process in order for the filed invention to receive a patent. In other words, from the perspective of patent examination, the rewritten claims are more refined than the original ones and can be used as training and evaluation data for the rewriting model.

We built an automatic patent claim rewriting model by fine-tuning a large language model (LLM) using the constructed dataset. Furthermore, we enhanced the performance of the automatic patent claim rewriting model by applying preference optimization based on the prediction results of patent examiners' Office Actions. We developed an automatic discrimination model to determine whether the claims generated by the model are likely to be accepted or rejected, taking into account the patent claims of patents cited by the examiner as prior art, and then used these prediction results for model optimization. In other words, we optimized the model to align with the preferences of patent examiners. We applied a method based on Kahneman-Tversky Optimization (KTO) for preference optimization. Our experimental results showed that our proposed model not only surpasses rule-based baseline models but also outperforms zero-shot learning in state-of-the-art LLMs. Furthermore, we demonstrated that preference optimization based on patent examiners' preferences significantly boosts the performance of patent claim rewriting.

Our contributions are as follows:
\begin{itemize}
\setlength{\parskip}{0cm}
\setlength{\itemsep}{0cm}
    \item We constructed a large-scale patent rewriting dataset of 2,245,640 cases by automatically extracting pairs of patents before and after examination, leveraging publicly available patent documents and amendment information.
    \item We developed an LLM-based patent rewriting model and explored several model tuning methods, including preference optimization based on patent examiners' preferences.
    \item  We showed that even small-scale models (0.5B parameters), when properly tuned with preference optimization, can outperform zero-shot learning in state-of-the-art LLMs, highlighting the importance of model tuning for this task.
\end{itemize}

Our ClaimBrush framework, which includes the dataset and rewriting models, will be available upon request and has the potential to significantly reduce the manual labor involved in refining patent claims and their evaluation in intellectual property practice.

\section{Related Work}

With the advent of large language models (LLMs) based on Transformer language models, methods using LLMs have shown state-of-the-art performance in various natural language processing tasks such as text summarization, question answering, machine translation, and grammatical error correction~\cite{vaswani2017attention,min2023recent}. While the performance of these applied tasks continues to evolve along with the improvement of LLMs, the development of patent language models specialized for the patent domain (especially U.S. patents), which have different vocabulary from general corpora, and their application to various tasks in the patent domain are progressing~\cite{lee2020patent,lee2023evaluating,tonguz2021automating}. In evaluating the performance of such patent language models, the focus is often on automatically generating patent claims. However, there is also the task that generates practical rewrites for given claim~\cite{shinmori2008multi}.

In the patent claim rewriting task, unlike simple generation tasks, it is necessary to generate rewrites that maintain the meaning content of the original claim, in other words, the essential information of the original invention, while improving its quality. Regarding the patent claim rewriting task, there is research based on rule-based methods that decompose a multi-multi claim (a claim that cites two or more other claims that selectively cite two or more other claims) into multiple claims and rewrite them. Nevertheless, this research is limited to rewriting support for very specific purposes~\cite{shinmorimultimulti}. More research must be done that automatically generates consistent rewrites, considering not only a single claim but also the entire structure of patent claims, for various purposes. However, existing publicly available datasets are insufficient for training and evaluating such rewriting generation models.

As research similar to our study, there is research on the proofreading task of scientific papers, which is approached from various aspects such as grammatical error correction and rewriting \cite{ito-etal-2019-diamonds,ito2020langsmith}, and review comment generation \cite{liang2023can,chamoun2024automated,kang2018dataset,lin2023moprd}. Typically, these studies utilize the papers submitted to sites such as OpenReview and their review comments for dataset construction. However, the pre-processing and annotation work of unstructured texts, such as papers and review comments, is costly, and there are still challenges in building models that correspond to various technical domains due to the limitations of data sources. On the other hand, patent documents are highly structured; their format is unified regardless of the domain, and they cover various technical domains. In addition, since patent documents contain rich meta-information and examination progress information, it may be possible to generate datasets for proofreading tasks across various technical domains automatically.

Given such background, as the first step toward realizing automatic proofreading models for technical documents, including patents, in this study, we automatically constructed a rewriting model for generating refined rewrites of patent claims and a training and evaluation dataset containing rich auxiliary information.

% done
\section{Dataset of Claim Refinement}

In order to train and evaluate rewriting generation models aimed at the automatic refinement of patent claims, it is necessary to prepare a reliable parallel corpus as a dataset that collects a large number of patent claim rewriting cases consisting of pairs of patent claims before and after rewriting. However, since patent claims are highly technical and legal documents, creating datasets manually is practically difficult from a cost perspective, as it requires workers to be well-versed in both the technical content and intellectual property practice.

In this paper, we collected pairs of patent claims described in the publication before examination and the patent publication associated with actual patent applications in Japan as rewriting examples for constructing such a dataset. In general, in patent examination, many patent applications are initially refused and then granted registration after applying necessary modifications (amendments) (see Fig.~\ref{flow}). In addition, minor amendments necessary for procedural purposes are also made regardless of whether the application is refused or not. The publication before examination is a patent document at the time of filing that is published after a certain period regardless of whether the patent is registered or not, and the patent publication is a patent document that has actually been granted registration. In Japan, these are known as 特許公開公報 (A) and 特許公報 (B9), and are available as open data. Considering the patent examination process, the patent claims in the patent publication have been modified as necessary from the patent document at the time of filing (corresponding to the publication before examination) in order to receive a patent. Thus, from the perspective of the patent system, the patent claims in the patent publication (=B9) can be considered as the result of "better rewriting" compared to those in the publication before examination (=A). Furthermore, we linked the examination history information published by the patent office to the collected rewriting pairs. In particular, we utilized the information on Office Actions, which are notifications of examination results by patent examiners (information on the legal provisions and prior patents that served as the basis for refusing the patent application).

Our approach to dataset construction is not limited to the published data of the Japan Patent Office but focuses on the internationally common patent examination system. Thus, the same approach can be applied to the published data of major patent offices worldwide, such as the USPTO and the EPO. However, due to differences in each country's patent laws and their implementation, building models tailored to each country's patent office is important. Japan, having the third-largest patent office in the world, presents unique characteristics that are valuable to model.

\begin{figure}[ht]
\centering
\includegraphics[width=0.85\columnwidth]{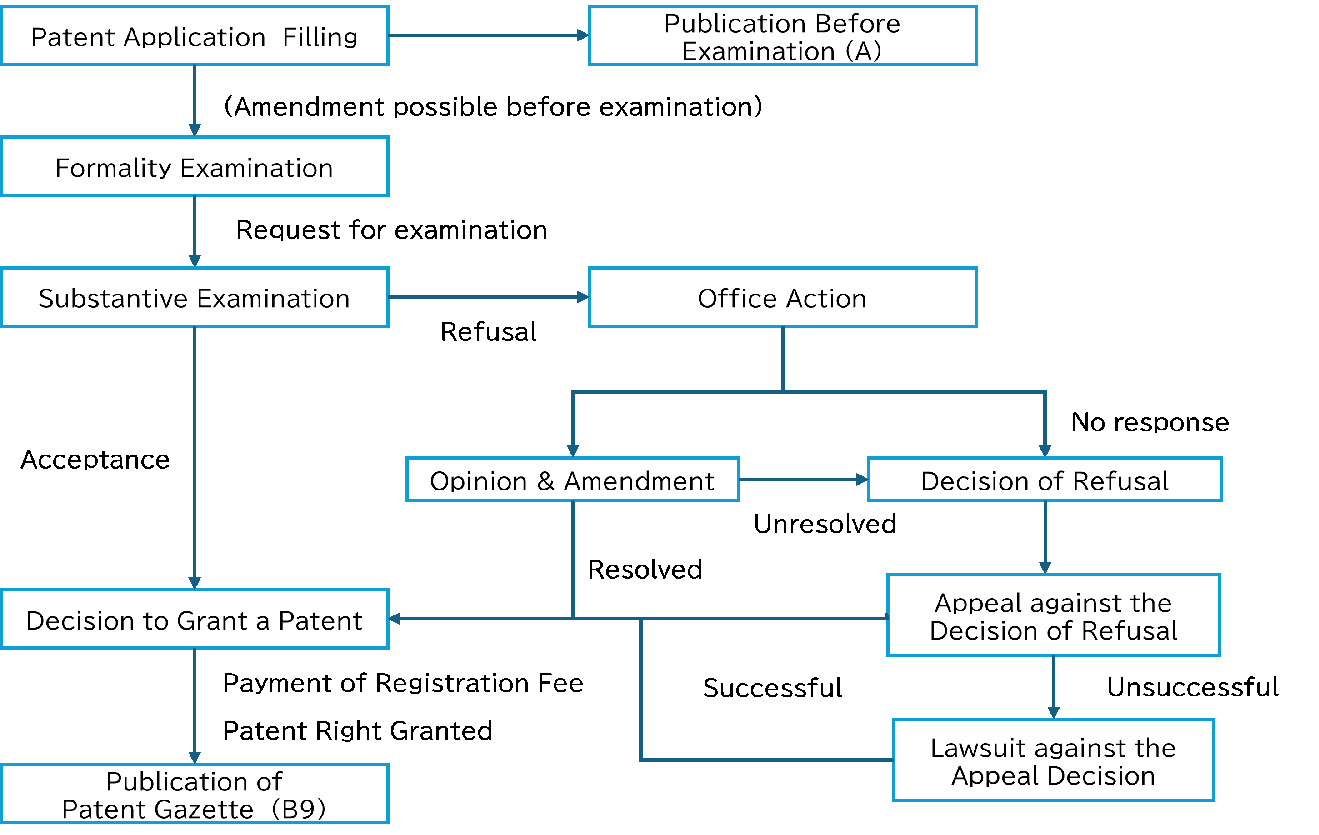}
\caption{Patent examination process in Japan.}
\label{flow}
\end{figure}

% done
\subsection{Dataset Construction}

We used the dump data of patent information provided by the Japan Patent Office for dataset construction\footnote{{\tiny https://www.jpo.go.jp/system/laws/sesaku/data/download.html}}. Specifically, from the set of publications before examination (特許公開公報(A)) and patent publications (特許公報(B9)) published through the Patent Office's system from 2004 to 2022, we extracted pairs of patent claims with the same patent application number. 
Furthermore, based on the patent application numbers, we linked the examination history information\footnote{{\tiny https://www.jpo.go.jp/system/laws/sesaku/data/keikajoho/index.html}} provided by the Patent Office and the IIP Patent Database~\cite{goto2007construction} to establish an integrated database available for patent information processing research. 

Among the additional information assigned to the rewriting pairs, we particularly utilized the legal provisions that served as the basis for refusing the patent application during examination\footnote{{\tiny https://www.japaneselawtranslation.go.jp/ja/laws/view/4097}}. We also utilized the claim information of prior patents cited by the patent examiner when notifying the reasons for refusal (the so-called Office Action). Currently, there are about 60 unique label numbers for the reasons for refusal\footnote{{\tiny https://www.j-platpat.inpit.go.jp/assets/pdf/C0710\_application.pdf}}, and multiple reasons for refusal may be assigned to a single patent application. For example, if the label for the reason for refusal is assigned as "Article 29, Paragraph 2," it suggests that the patent application may include issues with the inventive step. Similarly, patent examiners may cite multiple patents as prior art in an Office Action. An example of a rewriting pair and the main supplementary information used in the experiments in this research is shown in Table~\ref{tab:single_page}. Here, the legal provisions connected by ``$|$'' indicate that the actual reason for refusal violates any of those provisions, and the provisions connected by ``\textit{+}'' indicate that a combination of those legal provisions constitutes the reason for refusal. In some cases, multiple claims were consolidated into one, and the structure of the invention was made more specific to address the reasons for refusal cited by the patent examiner.
For example, Table~\ref{tab:single_page} shows an example of a rewriting pair where multiple claims were consolidated into one, and the structure of the invention was made more specific to resolve the reasons for refusal by the patent examiner. Such rewriting is known as a useful strategy for amending overly broad claims at the time of application.

\begin{table*}[!h]
\caption{Example rewriting case for patent application JP-A-2011-229635 (特開2011-229635)}

\centering
\small
\begin{tabular}{|c|p{5.8cm}|p{5.8cm}|}
\hline
\textbf{Section} & \textbf{Japanese} & \textbf{English Translation} \\
\hline
\textbf{Original Claims} & & \\
&【請求項1】 四角形状の板材の表面に所定の装飾シートが貼付されてなり、所定の遊技領域を有する遊技機用の遊技盤を構成する装飾板であって、

前記遊技領域の領域外に相当する部位でかつ前記板材の所定のコーナー部の近傍に相当する部位における前記装飾シートの樹脂表面層に対し、所定の情報を含んだ二次元コードよりなる識別コードがレーザー刻印されていることを特徴とする装飾板。

【請求項2】請求項1に記載の装飾板からなる遊技盤。

【請求項3】請求項2に記載の遊技盤を備えてなる遊技機。

& [Claim 1] A decorative board constituting a game board for a game machine having a predetermined game area, wherein a predetermined decorative sheet is affixed to a surface of a rectangular plate material, characterized in that an identification code consisting of a two-dimensional code containing predetermined information is laser-engraved on a resin surface layer of the decorative sheet at a portion corresponding to a vicinity of a predetermined corner portion of the plate material and corresponding to a portion outside the game area.

[Claim 2] A game board comprising the decorative board described in Claim 1.

[Claim 3] A game machine equipped with the game board described in Claim 2.
\\
\hline
\textbf{Amended Claims} & & \\
& 【請求項1】 四角形状の板材の表面に所定の装飾シートが貼付されてなり、所定の遊技領域を有する遊技盤であって、

遊技機の種別を特定可能な情報を含んだ第１識別コードが第１のコーナー部の近傍において付された四角形状をなす前記装飾シートを前記板材に貼付するシート貼付工程と、

前記板材に貼付された装飾シートにおける前記遊技領域の領域外に相当する部位でかつ前記第１のコーナー部と対角にあたる第２のコーナー部の近傍に相当する部位における前記装飾シートの樹脂表面層に対し、前記遊技機の種別を特定可能な情報を含んだ二次元コードよりなる第２識別コードをレーザー刻印により付す表面コード付け工程と、

前記コード付け工程を行うにあたり、前記第１識別コードの内容と前記第２識別コードの内容とを照合する照合工程と、

前記板材のうち、前記第２識別コードが付された部位を含む所定範囲を残して、前記第１識別コードが付された部位を切除する切除工程と、

前記板材の側面に対し、第３識別コードを付す側面コード付け工程とを経て形成された前記遊技盤を備えたことを特徴とする遊技機。

& [Claim 1] A game machine comprising a game board having a predetermined game area, wherein a predetermined decorative sheet is affixed to a surface of a rectangular plate material, characterized by including:
a sheet affixing step of affixing the rectangular decorative sheet, to which a first identification code including information capable of specifying a type of the game machine is attached in the vicinity of a first corner portion, to the plate material;
a surface code attaching step of attaching, by laser engraving, a second identification code consisting of a two-dimensional code including information capable of specifying the type of the game machine to a resin surface layer of the decorative sheet at a portion corresponding to a vicinity of a second corner portion diagonally opposite to the first corner portion and corresponding to a portion outside the game area of the decorative sheet affixed to the plate material;
a verification step of comparing contents of the first identification code and contents of the second identification code when performing the code attaching step;
a cutting step of cutting off a portion to which the first identification code is attached from the plate material while leaving a predetermined range including a portion to which the second identification code is attached; and
a side code attaching step of attaching a third identification code to a side surface of the plate material,
wherein the game board is formed through the above steps. \\
\hline

\textbf{Refusal Reasons} & 22:第29条第1項$|$第29条第2項$|$第29条第1項$+$第29条第2項 & 22: Article 29, Paragraph 1$|$Article 29, Paragraph 2$|$Article 29, Paragraph 1 $+$ Article 29, Paragraph 2 \\
\hline

\textbf{Prior Patent Numbers} & 特開2008-113853号公報 & JP-A-2008-113853 \\
& 特開2003-126393号公報 & JP-A-2003-126393 \\
& 特開2000-84166号公報 & JP-A-2000-84166 \\
& 特開2008-119277号公報 & JP-A-2008-119277 \\
& 特開昭60-21777号公報 & JP-A-1985-21777 \\
\hline
\end{tabular}
\label{tab:single_page}
\end{table*}

\begin{table*}[t]
    \centering
    \small
    \caption{Statistics of our dataset.}
    \begin{tabular}{l|c|c|c}
        \hline 
        Types & Freq. & Avg.chars & Avg.claims  \\ \hline \hline
         Type 1: A only (No corresponding B9) & 2,610,893 & 1402.16 & 7.84 \\ \hline 
         Type 2: A and B9 are identical (No reasons for refusal) & 369,807  & 1259.83 (+0.0\%) & 6.16 (+0.0\%) \\ \hline 
         Type 3: A and B9 are identical (With reasons for refusal) & 35,953  & 822.14 (+0.0\%) & 4.64 (+0.0\%) \\ \hline 
         Type 4: A and B9 differ (No reasons for refusal) & 628,882 & 1438.76 (-13.2\%) & 6.79 (-25.9\%) \\ \hline %
         Type 5: A and B9 differ (With reasons for refusal) & 1,210,998 & 1401.28 (-14.8\%) & 6.60 (-28.5\%) \\ \hline %
    \end{tabular}
    \label{tab:dataset}
\end{table*}

\subsection{Dataset Statistics}  
\label{sec:stat}

Table~\ref{tab:dataset} shows the statistics of valid pairs of patent claims between the publication before examination (A) and the patent publication (B9) included in our dataset. This includes cases where there are differences between the two, cases where there are no differences, cases with reasons for rejection, cases without reasons for rejection, and patent claims from the publication before examination (A) without a corresponding patent publication (B9) (Type 1). The average number of characters and claims are based on the patent claims in the patent publication (B9). The values in parentheses indicate the average percentage change compared to the publication before examination (A) prior to amendment. The percentage change for each application is calculated individually, and then averaged across all applications in each type. For ungranted patents (Type 1), the statistics are based on the publication before examination (A).

From Table~\ref{tab:dataset}, we can observe that the dataset includes text pairs of the publication before examination (A) and the patent publication (B9) for a total of 2,245,640 patent applications. Among these, there were 1,839,880 cases where there were differences in the text of the patent claims between the two (Types 4 and 5), and 405,760 cases where there were no differences (Types 2 and 3). Specifically, Type 4 (628,882 cases) represents applications where the texts differ but there are no reasons for rejection, indicating voluntary amendments made by applicants. Type 5 (1,210,998 cases) represents applications with differences and reasons for rejection, reflecting amendments made in response to Office Actions. We confirmed that the average number of characters and claims tends to decrease for patent publications compared to applications at the time of filing or publications before examination. Furthermore, when comparing the pairs with differences between filing and grant (Types 4 and 5), it is observed that the number of characters and claims in the patent publication decreases from the time of filing. The decrease in the number of claims is more pronounced in the amended pairs where reasons for rejection exist (Type 5) compared to those without reasons for rejection (Type 4). This indicates that patent applicants put significant effort into responding to Office Actions from patent examiners (Type 5), in contrast to the voluntary amendments made by applicants (Type 4). There are also cases where reasons for rejection were provided even though there were no differences between the amended pairs (Type 3). These are considered instances where the applicant's arguments were successful in overcoming the rejection without requiring amendment of the claims.

\section{Patent Claim Refinement Model Based on Large Language Model}

In this section, we define the task of patent claim refinement and describe the training approach of the patent claim rewriting generation model based on large language models (LLMs).

\subsection{Task Setting}

We define the patent claim refinement task as follows:
\begin{itemize}
\setlength{\parskip}{0cm}
\setlength{\itemsep}{0cm}

\item Input: Text $x = [c; r]$ concatenating the patent claims $c$ before rewriting and the additional information $r$. 

\item Output: Patent claims $c'$ after rewriting. 

\end{itemize}
Here, $c$ is the patent claims at the time of filing the patent application where refusal is expected, $c'$ is the patent claims to which rewriting has been applied so that it can receive a patent grant, and $r$ is the expected reason for refusal of patent application for $c$. Although multiple claims can be set for a single patent application, in this paper, we treat them together as a single text. In this task, given the input $x$, we generate $c'$ that maximizes the following probability using a language model:
{
\begin{equation}
{
c' = \mathop{\rm argmax}\limits_{c'} \prod_{t=1}^{T} P(c'_t|c'_{1:t}, x)
}
\end{equation}
}
Here, $c'_t$ is the $t$-th token of $c'$, $c'_{1:t}$ is the tokens up to the $(t-1)$-th token of $c'$, and $T$ is the number of tokens in $c'$. To solve this task, it is necessary to train a model with knowledge about rewriting patent claims.

\subsection{Supervised Fine-Tuning (SFT)}

To train the patent claim rewriting generation model, we applied Supervised Fine-Tuning (SFT) based on minimizing the following objective function to a pre-trained auto-regressive large language model\cite{radford2019language}:
{
\begin{equation}
\centering
L = -\sum_{(x, c') \in D} \sum_{t=1}^{T} \log P(c'_t|c'_{1:t}, x; \theta)
\end{equation}
}
Here, $D$ is the set of pairs of input text $x$ and rewritten patent claims $c'$ (training data), and $\theta$ is the set of trainable parameters of the model. For the patent claims used as input and output, we use those that have been granted registration after being refused once, among the pairs of publications before examination and patent publication included in our dataset. The template used for training the language model is as shown in Table~\ref{tab:template}. In the fine-tuning of the model, we test both full-parameter tuning and a method based on LoRA that learns only specific layers using low-rank adaptation of the original parameter matrix~\cite{hu2021lora}. 

\begin{table}[h]
    \centering
    \small
    \caption{Prompt template for patent claim rewriting.}
    %\begin{tcolorbox}%[colback=white!95!gray, colframe=black!75!white]
    \begin{description}
    \setlength{\parskip}{0cm}
    \setlength{\itemsep}{0cm}
    \scriptsize
    \item \texttt{<|im\_start|>system}
    \item \texttt{You are a helpful assistant.<|im\_end|>}
    \item \texttt{<|im\_start|>user}
    \item \texttt{Please rewrite the following patent claims, which may be refused, in a way that it can be published as a patent.}
    \item \texttt{\{input patent claims\}}
    \item \texttt{Expected refusal reasons:} 
    \item \texttt{\{list of refusal reasons\}
    \item <|im\_end|>}
    \item \texttt{<|im\_start|>assistant}
    \item \texttt{\{output patent claims\}}
    \item \texttt{<|im\_end|>}
    \end{description}
    %\end{tcolorbox}
    \label{tab:template}
\end{table}

\subsection{Preference Optimization}

We further refine the LLM fine-tuned by Supervised Fine-Tuning (SFT) for the proposed task using preference optimization techniques, which are known as a powerful way to strengthen language models based on human preferences. In this paper, we propose an approach using Kahneman-Tversky Optimization (KTO)~\cite{ethayarajh2024kto} based on patent examiners' preferences. As a conventional alignment approach for language models, Reward Learning from Human Feedback (RLHF) based on Proximal Policy Optimization (PPO) is considered a promising option~\cite{ouyang2022training,lee2023rlaif}. However, RLHF requires sequentially sampling generation results from the model during the learning process, which is computationally very expensive. Additionally, Direct Preference Optimization (DPO) requires pairs of desirable and undesirable responses~\cite{rafailov2024direct}. Similarly, KTO requires both desirable and undesirable responses, but they do not need to be in pairs. Due to the cost constraints of manual evaluation by patent examiners, we constructed a model that automatically estimates whether there are reasons to refuse the generated patent claims and used its feedback results for preference optimization.

\subsection{Kahneman-Tversky Optimization}

KTO is a preference optimization method based on prospect theory that incorporates human decision-making biases into the loss function. The Kahneman-Tversky Optimization (KTO) process directly maximizes the utility of generations instead of maximizing the log-likelihood of the preferences. KTO only requires a binary signal of whether output is desirable or not, which is a kind of data easier to obtain than paired preference data. The loss function of KTO is defined as follows:
\begin{equation}
L_{\text{KTO}}(\pi_\theta, \pi_{\text{ref}}) = \mathbb{E}_{(x,y) \sim D}[\lambda_y - v(x,y)]
\end{equation}
Here, $\lambda_y$ is a weighting term that becomes $\lambda_D$ ($\lambda_U$) when $y$ is a desirable (undesirable) response, and $v(x,y)$ is a function defined as follows:
{\small
\begin{equation}
v(x,y) = \left\{
\begin{array}{ll}
\lambda_D \sigma(\beta(r_\theta(x,y) - z_0)) & , y \sim y_{\text{desirable}} \\
\lambda_U \sigma(\beta(z_0 - r_\theta(x,y))) & , y \sim y_{\text{undesirable}}
\end{array}
\right.
\end{equation}
\begin{equation}
r_\theta(x,y) = \log \frac{\pi_\theta(y|x)}{\pi_{ref}(y|x)}
\end{equation}
\begin{equation}
z_0 = \mathbb{E}_{(x',y') \sim D} \mathrm{KL} ( \pi_{\theta}(y'|x') \parallel \pi_{\text{ref}}(y'|x'))
\end{equation}
}
Here, $x$ is the input to the language model, $y=c'$ is the response of the model, $\pi_\theta$ is the language model to be learned, $\pi_{ref}$ is the reference model with fixed parameters, $\sigma$ is the sigmoid function, and $\beta > 0$ is a hyper-parameter that controls the degree of risk aversion. In other words, in KTO, when $y$ is a desirable (undesirable) response, the loss becomes smaller by increasing (decreasing) $r_\theta(x,y)$. On the other hand, if the change in $r_\theta(x,y)$ is too large, the reference point $z_0$ also becomes large, offsetting the decrease in loss. This allows the model to accurately learn the desirability of responses while suppressing deviation from $\pi_{ref}$. The weighting terms of KTO are determined considering the imbalance in the number of desirable and undesirable responses.

\subsubsection{Automatic Feedback From Preference Model}

To model patent examiners' preferences, we fine-tuned a pre-trained LLM for binary classification using our dataset of examples where patent claims were rewritten in response to reasons for refusal. Specifically, we consider two versions of patent claims: \(c\), the claims from the publication before examination, and \(c'\), the claims from the patent publication after amendments. We construct a prompt text by concatenating the patent claims \(\hat{c} \in \{ c, c' \}\) to be evaluated, where \(\hat{c}\) represents either the original claims \(c\) or the amended claims \(c'\), the given reason for refusal \(r\), and the claims \(a\) of a prior patent related to \(\hat{c}\). When there are multiple prior patents cited, we randomly sample one to use as \(a\) to simplify the input and reduce computational complexity. In training, the claims \(\hat{c}\) are either taken from the publication before examination \(c\) (undesirable examples that failed to overcome the refusal) or from the patent publication \(c'\) (desirable examples that successfully overcame the refusal). By learning from these past successful and unsuccessful cases, the model learns to judge whether a set of claims can avoid refusal (i.e., whether they can be granted a patent), effectively capturing patent examiners' preferences. The preference model is formulated as follows:

\begin{equation}
p_{\phi}(y|\hat{c}, r, a) = \sigma(\mathbf{w}^{\mathrm{T}} \mathbf{h} + b)
\end{equation}
Here, $\mathbf{h} = f_{\text{LM}}(\hat{c}, r, a)$ is the output of the final layer of the pre-trained language model, $\mathbf{w}, b$ are trainable parameters, and $\sigma$ is the sigmoid function. The cross-entropy error is used for training the preference model. 

In the training process of KTO using this preference model, the desirability of the generated rewriting $\hat{c}$ corresponding to the original claims $c$ by the rewriting model is determined based on:
\begin{equation}
y = \begin{cases}
y_{\text{desirable}} & , p_{\phi}(y|\hat{c}, r, a) \geq 0.5 \\
y_{\text{undesirable}} & , p_{\phi}(y|\hat{c}, r, a) < 0.5.
\end{cases}
\end{equation}

This is expected to automatically construct the preference data necessary for KTO learning from the preference model without requiring manual annotation. The modeled preferences of patent examiners can potentially be applied to automate patent examination and automatically evaluate the quality of patents. Thus, we not only utilize the constructed preference model of patent examiners to fine-tune the rewriting model based on LLMs but also apply it to the rewritten results generated by the rewriting model to evaluate whether the rewriting would actually be accepted as a patent. This enables us to assess the performance of the rewriting model in the patent claim refinement task from a more practical perspective, providing valuable insights into the model's effectiveness in generating patent-worthy claims.

\section{Experiment Settings}

In this section, we describe the experimental settings for evaluating the patent claim refinement performance of the proposed model.

\subsection{Dataset}

Among the rewriting pairs included in the dataset we constructed, we used pairs where the patent was able to be granted after being refused once for some reason during the examination of the patent application. However, since training including pairs with very long sentences requires a certain amount of computational resources, we constructed data for training (317,182 cases), validation (1,000 cases), and evaluation (1,000 cases) of language models and the preference model using only pairs with not too long contexts\footnote{{\footnotesize We used samples where the token length of the concatenated input and output text, tokenized by the tokenizer of Qwen1.5, was 2400 or less. }}. In addition, for preference optimization, we prepared 1,000 samples that do not overlap with these data.

\subsection{Comparison Models}
We compared the performance of heuristic baseline models, zero-shot learning in state-of-the-art large language models (LLMs), and proposed models based on fine-tuning of LLMs. The models for comparison are as follows:
\begin{itemize}
\setlength{\parskip}{0cm}
\setlength{\itemsep}{0cm}

\item \textbf{Copy}: The case where the patent claims are not rewritten.
\item \textbf{RDC} (Random Delete of Claims): A case where one of the claims, excluding claim 1, is randomly deleted. Any claims dependent on the deleted claim are also excluded. This is based on the observation that the number of claims tends to decrease from the time of application to patent registration (see Section~\ref{sec:stat}).
\item \textbf{DMMC} (Delete of Multi-Multi Claims): A case where all multi-multi claims are excluded. Similarly, any claims dependent on the deleted multi-multi claims are also excluded. This is based on the fact that citing multi claims in the form of multi-multi claims is prohibited or discouraged by patent offices in various countries\footnote{{\tiny https://www.jpo.go.jp/e/system/patent/shinsa/multimulticlaims.html}}.
\item \textbf{LLM} (Zero-shot): We evaluated the zero-shot performance of state-of-the-art LLMs. While these LLMs are not specifically designed for patent claim refinement, they have shown the ability to solve various tasks in a zero-shot setting~\cite{li2023practical}. These models were included as baselines to assess their performance on the patent claim refinement task without fine-tuning. The input format provided to the models was the same as in the fine-tuning experiments, including the patent claims and expected reasons for rejection, and the models were asked to generate refined patent claims without additional training.
\item \textbf{LLM} (SFT): The case where the LLM is fine-tuned using full-parameter supervised fine-tuning.
\item \textbf{LLM} (LoRA): The case where LoRA (Low-Rank Adaptation) is applied to the LLM, and only specific layers are fine-tuned.
\item \textbf{LLM} (KTO): The case where the supervised fine-tuned LLM is further optimized using preference optimization via Kahneman-Tversky Optimization (KTO).
\end{itemize}

For the parameter size $|\theta|$ of the LLM used for fine-tuning, we used Qwen1.5\footnote{{\tiny https://huggingface.co/Qwen}} with parameter sizes of {0.5B, 1.8B, 7B}, considering both learning cost and inference speed. To fair comparison, fine-tuning was performed within the same model family. We tested full-parameter tuning as well as LoRA, which adapts the low-rank approximation of the original parameter matrix. Due to computational constraints, we only used SFT for models with 0.5B parameters. The Qwen1.5-0.5B model was further enhanced using KTO after SFT. For the preference model, we added a linear layer to the final layer of the supervised fine-tuned Qwen1.5-0.5B model and trained it for binary classification\footnote{\tiny{https://huggingface.co/docs/transformers/en/model\_doc/qwen2}}. For zero-shot learning, we used GPT-4o-mini, GPT-3.5-turbo, and Qwen-1.5-72B, the largest model in the Qwen family. Qwen-1.5-72B significantly outperforms GPT-3.5-turbo while slightly trailing behind GPT-4 in general task performance.

\subsection{Training Setting}
In the SFT of the language model and the training of the preference model, we set the learning batch size to 64 and the model's Optimizer to AdamW with a learning rate of 5e-5. The number of learning epochs for the model was set to 1, and the learning rate was linearly decayed to zero at the end. Also, in the training of the preference model, we set the learning batch size to 64 and the model's Optimizer to AdamW with a learning rate of 5e-5. In KTO, we set the learning batch size to 64 and the model's Optimizer to AdamW with a learning rate of 1e-5. The number of learning epochs for the model was set to 3, and a warmup of 200 steps was applied, after which the learning rate was linearly decayed to zero at the end. In generating the training samples for KTO, we randomly generated 12 rewriting candidates (top\_p=0.95, temperature=0.7) for each sample from the training dataset and automatically assigned the labels of desirable and undesirable based on the evaluation from the preference model. Here, we added the case of outputting the input text as is as undesirable and the correct rewrite as desirable. Also, we set $\lambda_D$ to 3.0, $\lambda_U$ to 1.0, and $\beta$ to 0.2. In LoRA training, we set $r=8$ and $\alpha=16$ for the linear layers in the self-attention layer of the model.

% 小数点2桁まで算出する
\begin{table}
    \centering
    \small
    \caption{Prediction performance of preference model.}
    \begin{tabular}{c|c|c|c|c}
        \hline 
        Labels & Prec. [\%] & Recall [\%] & F1 [\%] & Freq.\\ \hline \hline
        Undesirable & 71.81 & 77.70 & 74.64 & 1000 \\ \hline 
        Desirable & 75.71 & 69.50 & 72.47 & 1000 \\ \hline 
    \end{tabular}
    \label{tab:pref}
\end{table}

% seedを固定する
\begin{table*}
    \centering
    \small
    \caption{Performance of patent claim refinement.}
    \begin{tabular}{l|c|c|c|c|c|c} \hline 
      Models   & $|\theta|$ &  GLEU$\uparrow$(word) & GLEU$\uparrow$(phrase) & SARI$\uparrow$(word) & SARI$\uparrow$(phrase) & Accept. Rate$\uparrow$ \\ \hline \hline 
       Reference & - & 100.0 & 100.0  & 100.0 & 100.0 & 0.69 \\ \hline 
       Copy & - & 63.23 & 23.08  & 28.38& 20.21 & 0.22 \\ \hline 
       RDC  & - & 47.22& 18.15 & 28.38& 20.21 & 0.30 \\ \hline 
       DMMC  & - &  57.85& 21.89 & 28.38& 20.21 & 0.23  \\ \hline 
       gpt4o-mini (zero-shot)   & -    &  47.04 & 5.17  & 32.00 & 24.99 & 0.70\\ \hline 
       gpt3.5-turbo (zero-shot) & -    &  40.19 & 11.75 & 29.62 & 27.33 & 0.17  \\ \hline 
       Qwen1.5-72B (zero-shot)  & 72B  &  29.97 & 1.45 & 27.23 & 20.94 & 0.69 \\ \hline 
       Qwen1.5-0.5B-LoRA       & 0.5B & 54.16  & 22.17 & 34.55    & 37.83 & 0.53\\ \hline 
       Qwen1.5-0.5B-LoRA  w/o $r$ & 0.5B & 32.47 & 11.40 & 34.19  & 37.36 & 0.53\\ \hline 
       Qwen1.5-1.8B-LoRA & 1.8B & 41.29 & 15.24  & 34.22 & 37.37  & 0.49  \\ \hline 
       Qwen1.5B-7B-LoRA  & 7B   & 55.67  & 24.10 & 35.42  & 39.12 & 0.57 \\ \hline 
       Qwen1.5-0.5B-SFT  & 0.5B & 55.06 & 24.65  & 36.66 & 40.54  & 0.63 \\ \hline 
       Qwen1.5-0.5B-SFT w/o $r$ & 0.5B  & 57.25  & 25.13 & 36.72  & 40.39 & 0.60 \\ \hline
       Qwen1.5-0.5B-KTO  & 0.5B & \bf 58.45 & \bf 25.76 & \bf 38.95 & \bf 43.92 & \bf 0.96 \\ \hline 
       
    \end{tabular}
    \label{tab:results}
\end{table*}

\subsection{Evaluation Metrics}

The patent claim rewriting task is essentially a machine translation task within the same language. However, considering the amendment requirement provisions for patents stipulated in Article 17 of the Patent Act, evaluation metrics that take into account the input text itself are necessary. Therefore, we adopted standard metrics used in the fields of grammatical error correction and text simplification~\cite{wang2021comprehensive,al2021automated}, where consideration of not only the reference and hypothesis but also the input is necessary. Furthermore, using the preference model trained for KTO, we determine the appropriateness of the output patent claims of the rewriting model. The evaluation metrics used in this research are as follows:
\begin{itemize}
\setlength{\parskip}{0cm} 
\setlength{\itemsep}{0cm} 
 
\item \textbf{GLEU}: An evaluation metric that improves BLEU, a standard evaluation metric for machine translation, for the task of grammatical error correction~\cite{napoles2015ground}. GLEU is calculated by subtracting the number of n-grams that appear in the original text but not in the reference text from the number of n-grams that match between the corrected text and the reference text.

\item \textbf{SARI}: An evaluation metric in the field of text simplification~\cite{xu2016optimizing}. SARI is calculated by comparing three texts: the input text, the output text, and the correct text. Specifically, it is calculated by the average of the F-score of the n-grams correctly added to the input text by the model, the F-score of the n-grams correctly retained from the input text, and the precision of the n-grams correctly deleted from the input text.

\item \textbf{Acceptance rate (Accept. Rate)}: This metric evaluates whether the rewritten patent claims generated by the model are likely to be approved or refused by the patent examiner, taking into account prior patents. Instead of the patent examiner, we use the preference model trained for KTO. The model evaluates each set of claims and assigns a label indicating whether the claims are likely to be accepted (desirable) or rejected (undesirable). The acceptance rate is calculated as the proportion of generated claims that are predicted to be desirable by the preference model.

\end{itemize}
For the calculation of GLEU and SARI, we adopt both word-level segmentation and phrase-level text segmentation results considering the composition of the invention.

% UpdateROUGE, 

\section{Experimental Results}

\subsection{Performance of preference prediction}

Table~\ref{tab:pref} shows the prediction results of the preference model for patent examiners' in the evaluation set. The results show that both the prediction of desirable and undesirable achieved an F1 score of over 73\%. Since the frequency of each label is uniform, it can be seen that it significantly exceeds the chance rate accuracy of 50\%. In other words, the preference prediction model can predict with high confidence whether the rewritten patent claims will be judged as acceptable or unacceptable by the examiner against the claims of prior patent, suggesting the possibility of substituting patent examiners' judgment of the appropriateness of patent claims by automatic evaluation.

\subsection{Performance of patent claim refinement}

Table \ref{tab:results} shows the performance of each patent claim rewriting generation model in the evaluation set. Here, w/o $r$ denotes the case where the reasons for patent refusal are not used as input to the model. The proposed method based on LLMs surpasses the heuristics baseline models in SARI (word, phrase) using word-level and phrase-level segmentation criteria. The Copy model, which adopts the original patent claims as output, outperforms the proposed method based on LLMs in GLEU (word) using word-level segmentation criteria. On the other hand, RDC based on random deletion of claims shows lower GLEU than the method based on LLMs, except for the LoRA model based on low-parameter-size LLMs. DMMC based on deletion of multi-multi claims resulted in competing results with Copy because there were only a few multi-multi claims in the evaluation set results, but the GLEU decreased from copy. This reflects the characteristic of GLEU that overestimates copying from the input text and underestimates the bold strategy of deleting long sentences. On the other hand, for SARI, which reflects each of the rewriting operations of addition, deletion, and substitution in the evaluation, Copy and RDC show low performance. Also, the acceptance rate, which is the automatic evaluation result of the refusal judgment of the generated patent claims, never exceeded 50\% for the baseline models, in contrast to the method based on LLMs.

Regarding the effect of different learning approaches in the training of the rewriting model, SFT outperformed LoRA. Due to the nature of the LoRA learning method, it strongly depends on the performance of the language model used as the base. The language model used in this experiment is a multilingual LLM, but there is room for improvement in its ability to generate Japanese patent documents.
when using LoRA learning, a 7B model shows results that finally compete with the SFT of a 0.5B model, suggesting the importance of adapting the pre-trained model to both Japanese and the patent domain. Also, the model to which KTO was applied achieved a 96\% acceptance rate of the rewriting results of patent claims by aligning with the preference model of patent examiners' judgment of patent refusal, showing a 36\% improvement from the SFT model used as the base in the KTO learning. Similarly, the evaluation of GLEU and SARI also shows improvement from the SFT model, suggesting the possibility of applying it to patent examiners' preferences in patent examination while at least maintaining the patent claim rewriting performance. Regarding the effect of utilizing the information of reasons for refusal as input to the model, it showed a certain effect when applying LoRA learning, but was limited when applying SFT. This result suggests the need to provide further input as a clue to resolve the reasons for refusal, rather than just providing the reasons for refusal.

Regarding zero-shot learning, GPT-4o-mini, GPT-3.5-turbo, and Qwen-1.5-72B significantly underperformed compared to smaller, fine-tuned models, often even falling short of weaker baseline models like RDC and DMMC. These larger models particularly struggled with phrase-level evaluation, likely due to their tendency to produce disruptive rewrites that violated the amendment requirements of Article 17 of the Patent Act. Additionally, GPT-3.5-turbo and Qwen-1.5-72B frequently generated outputs that deviated from the conventions of Japanese patent language, at times resembling direct translations from English patents. In contrast, the smaller, fine-tuned models consistently outperformed the much larger LLMs, highlighting the importance of domain-specific adaptation in specialized fields such as Japanese patents.

%\begin{table}[t]
%    \centering
%    \tiny
%    \caption{Statistics of generated writings.}
%    \begin{tabular}{l|c|c|c|c}
%        \hline 
%        Types & Avg.chars & Avg.claims \\ \hline \hline
%         aaaa & 1402.16 & 7.84 && \\ \hline 
%    \end{tabular}
%    \label{tab:resultanalysis}
%\end{table}

\section{Conclusion}
In this paper, we proposed ClaimBrush, a novel framework that includes a rewriting model and a dataset for training and evaluation, aimed at automatic patent claim refinement. We constructed an automatic patent claim rewriting model by fine-tuning a large language model using the constructed dataset. Furthermore, we enhanced the performance of the automatic patent claim rewriting model by applying preference optimization based on the prediction results of patent examiners' Office Actions. The experimental results showed that our proposed model not only surpasses heuristic-based baseline models but also significantly outperforms zero-shot learning in state-of-the-art large language models. Furthermore, we demonstrated that preference optimization based on patent examiners' preferences boosts the performance of patent claim rewriting.

In terms of future work, efforts should focus on the improvement of the model’s efficiency in handling the long and complex nature of patent claims, potentially through the integration of more efficient architectures such as encoder networks. Additionally, the adaptation of pre-trained models to the patent domain, particularly for Japanese patent-related tasks, remains crucial for achieving better performance. The exploration of advanced methods such as Retrieval-Augmented Generation (RAG)~\cite{lewis2020retrieval} to incorporate external information from prior arts and patent body text could offer further improvements. Finally, the development of more accurate evaluation metrics that align with human judgments is essential for refining the effectiveness of patent claim rewriting.

\section*{Acknowledgement}
This work was supported by JSPS KAKENHI 22K17958 and JST ACT-X JPMJAX22A4.

\bibliographystyle{IEEEbib}
\bibliography{custom}

\clearpage
\label{sec:refs}

\end{document}